\documentclass{article}
\usepackage{amssymb}
\usepackage{amsmath, amssymb}
\usepackage{graphicx}
\usepackage{multicol}
\usepackage[bookmarks=true]{hyperref}
\usepackage{wrapfig}
\usepackage{fancyvrb} %
\usepackage{graphics} %
\usepackage{epsfig} %
\usepackage{times} %
\usepackage{amsmath} %
\usepackage{amssymb}  %
\usepackage{siunitx} %
\usepackage{textcomp}
\usepackage{gensymb} %
\usepackage{booktabs}
\usepackage{multirow}
\usepackage{enumitem}
\usepackage{xspace}
\usepackage{makecell}
\usepackage{multicol}
\usepackage{rotating}
\usepackage{contour}
\usepackage{courier}
\usepackage{subcaption}
\usepackage[font=small]{caption}
\usepackage[percent]{overpic}
\usepackage[11pt]{moresize}
\usepackage[dvipsnames]{xcolor}
\usepackage{booktabs}
\usepackage{siunitx}

\usepackage{ltablex}
\usepackage{csquotes}
\usepackage{tcolorbox}
\usepackage{fancyvrb}
\tcbuselibrary{skins} %
\usepackage{newfloat}

\usepackage[normalem]{ulem}
\usepackage[preprint]{corl_2025} %

\newcommand{\methodname}[1]{{\textsc{Casper}\xspace}}

\newcommand{\teleop}[1]{{Full Teleop#1}}

\newcommand{\heuristic}[1]{{HAT#1}}
\newcommand{\heuristicbayesian}[1]{{RBII#1}}

\definecolor{mycitecolor}{HTML}{8F00FF}  
\hypersetup{
    colorlinks,
    citecolor=mycitecolor,
    filecolor=black,
    linkcolor=mycitecolor,
    urlcolor=mycitecolor
}

\tcbuselibrary{breakable}       

\newtcolorbox{participantquote}{%
  breakable,                    
  colback=violet!5,               
  colframe=violet!50,             
  boxrule=0.4pt,                
  left=4pt,right=4pt,
  top=4pt,bottom=4pt,
  enhanced,                     
  fonttitle=\bfseries,          
}

\usepackage{listings}
\title{\methodname{}: Inferring Diverse Intents for Assistive Teleoperation with Vision Language Models}

\author{
Huihan Liu$^{1}$, 
Rutav Shah$^{1}$,
Shuijing Liu$^{1}$, Jack Pittenger$^{1}$, Mingyo Seo$^{1}$, \\[5pt]
\textbf{Yuchen Cui$^{2}$, Yonatan Bisk$^{3}$, Roberto Mart{\'i}n-Mart{\'i}n$^{1}$, Yuke Zhu$^{1}$}
\\[5pt]
$^1$The University of Texas at Austin ~ $^2$The University of California, Los Angeles \\ $^3$ Carnegie Mellon University
}

\begin{document}
\maketitle

\vspace{-2mm}
\begin{abstract} 
Assistive teleoperation, where control is shared between a human and a robot, enables efficient and intuitive human-robot collaboration in diverse and unstructured environments.
A central challenge in real-world assistive teleoperation is for the robot to infer a wide range of human intentions from user control inputs and to assist users with correct actions. Existing methods are either confined to simple, predefined scenarios or restricted to task-specific data distributions at training, limiting their support for real-world assistance.
We introduce \methodname{}, an assistive teleoperation system that leverages commonsense knowledge embedded in pre-trained visual language models (VLMs) for real-time intent inference and flexible skill execution. 
\methodname{} incorporates an open-world perception module for a generalized understanding of novel objects and scenes, a VLM-powered intent inference mechanism that leverages commonsense reasoning to interpret snippets of teleoperated user input, and a skill library that expands the scope of prior assistive teleoperation systems to support diverse, long-horizon mobile manipulation tasks. 
Extensive empirical evaluation, including human studies and system ablations, demonstrates that \methodname{} improves task performance, reduces human cognitive load, and achieves higher user satisfaction than direct teleoperation and assistive teleoperation baselines. More information is available at \url{https://ut-austin-rpl.github.io/casper/}

\end{abstract}


\section{Introduction}

Deploying robots in human-centric settings like households requires balancing robot autonomy with humans' sense of agency \cite{electronics11193065, Mostafa2019-MOSAAA-6, level_autonomy, collier2025senseagencyassistiverobotics, loehr2022sense, wen2019sense}. Full teleoperation offers users fine-grained control but imposes a high cognitive load, whereas fully autonomous robots act independently but often misalign their actions with nuanced human needs. \textbf{Assistive teleoperation} --- a paradigm in which both the human and the robot share control \cite{Dragan2013APF, dragan2012formalizing, argall2018autonomy, chen2022ashaassistiveteleoperationhumanintheloop} --- has thus emerged as an ideal middle ground. By keeping the user in control of high-level decisions while delegating low-level actions to the autonomous robot, this approach both preserves user agency and enhances overall system performance. As such, assistive teleoperation is becoming a desirable paradigm for robots to serve as reliable partners in human-centric environments, such as assisting individuals with motor impairments \cite{Brose2010TheRO, Miller1998AssistiveRA}.

\begin{figure}[t]
    \centering
      \includegraphics[width=1\linewidth]{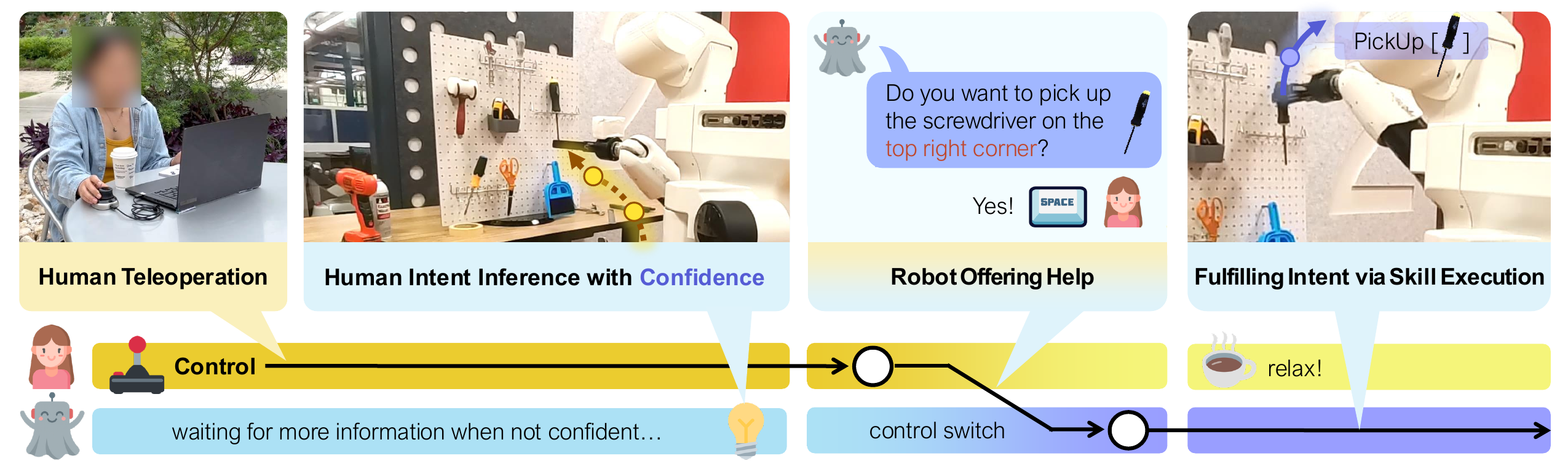}
		\captionof{figure}{\textbf{\methodname{} infers user intents and offers help when confident.} Given user teleoperation input, \methodname{} uses VLMs to predict human intent using commonsense reasoning. Upon user confirmation, \methodname{} performs autonomous execution to fulfill the intent using a skill library. \methodname{}'s background reasoning runs in parallel with foreground human control to minimize disruption.\label{fig:teaser}
        } 
        \vspace{-4mm}
\end{figure}

While promising, assistive teleoperation in everyday environments remains challenging. A long-standing challenge in assistive teleoperation is to infer human intents from user control inputs and assist users with correct actions~\cite{dragan2012formalizing}.
This challenge is amplified in real-world settings, where robots must go beyond closed-set intent prediction \cite{belsare2025zeroshotuserintentrecognition, zhao2024conformalizedteleoperationconfidentlymapping} to handle diverse, open-ended user goals across different contexts and scenes.
As a result, a key capability the robot should possess is to interpret user control inputs within the visual context and infer intent through commonsense reasoning. 
For example, consider a user teleoperating a robot to move a jar of pasta toward both a laptop and a cooking pot. Even if the pasta jar is closer to the laptop, commonsense suggests that the user intends to pour pasta into the pot, not onto the laptop. 
As another example, some users push an automatic door to open it, while others want to press an accessibility button. 
These examples illustrate the nuanced and context-dependent nature of human intent, highlighting the level of commonsense reasoning required for robots to provide effective and satisfactory assistance.

Existing assistive teleoperation systems often fall short in inferring diverse intents. Prior methods often limit the problem space to a closed set of objects \cite{zhao2024conformalizedteleoperationconfidentlymapping, argall2018autonomy}, or to a predefined task like picking up objects, implicitly assuming the intent type is known a priori \cite{zhao2024conformalizedteleoperationconfidentlymapping, belsare2025zeroshotuserintentrecognition}. 
These intent inference methods, either based on rule-driven strategies \cite{Padmanabha_2024, belsare2025zeroshotuserintentrecognition} or learned from demonstrations \cite{Cui_2023,pmlr-v164-karamcheti22a,zhao2024conformalizedteleoperationconfidentlymapping,chen2022ashaassistiveteleoperationhumanintheloop}, are typically limited to one single skill type or bound by the task distributions at training, struggling to generalize in new scenarios. Critically, these systems usually lack commonsense reasoning, which is essential for interpreting contextual cues and generalizing intent inference to novel scenes and behaviors in real-world environments.

To address the above limitations, we introduce \methodname{}, an assistive teleoperation system that infers diverse intents from human user control and offers assistance with long-horizon mobile manipulation tasks (Fig.~\ref{fig:teaser}).
\methodname{} builds on three core components.
First, it features an open-world perception module that uses pre-trained visual language models (VLMs) to provide a generalized understanding of open-world objects and scenes without task-specific training.
Second, \methodname{} leverages VLM-powered commonsense reasoning to infer a diverse range of user intents, significantly expanding the possible intent choices compared with prior systems.
Third, to realize task execution, \methodname{} uses a flexible library of parameterized skills encompassing a range of navigation and contact-rich manipulation behaviors~\cite{shah2024bumbleunifyingreasoningacting}. 
With this comprehensive and composable skill library, \methodname{} can execute long-horizon tasks that go beyond the capabilities of traditional assistive teleoperation systems.

Furthermore, deploying the system for long-horizon tasks introduces a user-centric consideration: offering undesirable assistance based on prematureintent inference can frustrate or disrupt the user. To avoid this, the system should determine intents only after gathering enough information from user inputs and visual contexts. 
\methodname{} addresses this by shadowing the user: it observes foreground human actions and infers user intents in the background.
A confidence module based on self-consistency \cite{wang2023selfconsistencyimproveschainthought} ensures that assistance is triggered only when prediction confidence is high, reducing errors and user disruption.
By running VLM-based inference in parallel with user control, \methodname{} unobtrusively predicts intent and prepares actions. 

To evaluate the effectiveness of \methodname{} in assisting human users, we conduct extensive user studies on a mobile manipulator (TIAGo \cite{pal-tiago}), involving $10$ pilot study participants and $13$ study participants, totaling over $80$ hours of interaction across $3$ long-horizon tasks. Additionally, we conduct offline experiments to test the intent inference module and perform detailed performance analyses and ablation studies.
Compared with prior assistive teleoperation baselines without commonsense reasoning ability and a full teleoperation baseline, \methodname{} achieves a higher success rate, better user satisfaction, and lower cognitive load of users across all tasks.

\section{Related Work}

\textbf{Assistive Teleoperation.} 
Assistive teleoperation offers a promising balance between human control and robotic assistance, enhancing user agency and task efficiency~\cite{Brose2010TheRO, Miller1998AssistiveRA, Padmanabha_2024, Padmanabha_2024_voice, jenamani2024flairfeedinglonghorizonacquisition,liu2024dragon}.
Assistive teleoperation enables users to share control with the robot, injecting their intent to guide the system toward their goals~\cite{dragan2012formalizing, Dragan2013APF, chen2022ashaassistiveteleoperationhumanintheloop, pmlr-v144-karamcheti21a, You2011AssistedTS, Broad2017LearningMF, Javdani2017SharedAV}.
Accurately predicting user intent is thus a key challenge~\cite{dragan2012formalizing, Gopinath2020ActiveID, hoffman2024inferring}.
Prior approaches typically select the most probable intent from a fixed set of goals~\cite{dragan2012formalizing, Admoni2016PredictingUI, Balanced8673192, Nikolaidis_2017, newman2020harmonicmultimodaldatasetassistive}, assume a single predefined skill~\cite{belsare2025zeroshotuserintentrecognition}, or use data-driven methods to map high-dimensional user inputs to low-dimensional actions within specific tasks~\cite{chen2022ashaassistiveteleoperationhumanintheloop, zhao2024conformalizedteleoperationconfidentlymapping, Cui_2023, pmlr-v164-karamcheti22a, pmlr-v144-karamcheti21a, yoneda2023noisebackdiffusionshared, Jonnavittula2022LearningTS, Zurek2021SituationalCA, Schaff2020ResidualPL}.
However, both approaches struggle to generalize beyond predefined intents without retraining or reprogramming. Moreover, they also lack the commonsense reasoning capability to interpret human control input within the visual context.

\textbf{Human Intent Inference.}
Inferring hidden human states is a critical step toward understanding human behavior for a wide range of downstream tasks~\cite{lai2024humanactionanticipationsurvey, mascaro2024intentionconditionedlongtermhumanegocentric, girase2021lokilongtermkey, PIE9008118, liu2020spatiotemporalrelationshipreasoningpedestrian}.
In robotics, intent inference enables robots to operate effectively in human-centered environments~\cite{huang2024litlargelanguagemodel, ali2024comparingapplesorangesllmpowered, huang2023hierarchicalintentiontrackingrobust, liu2023intention}.
To achieve shared goals, robots must reason about a human collaborator’s latent strategy~\cite{wang2021co, liu2021learning}, future actions~\cite{liu2023intention, wang2024mosaic, huang2024litlargelanguagemodel}, goals~\cite{chang2020robot, huang2023hierarchicalintentiontrackingrobust, liu2024dragon}, and preferences~\cite{wu2023tidybot, wang2024apricot} to adjust their behavior accordingly. \methodname{} advances these efforts by leveraging VLM-based intent inference to facilitate assistance in assistive teleoperation settings.

\textbf{LLMs and VLMs for Robotics.}
Foundation models, pretrained on internet-scale data, have gained attention for their strong generalization and adaptability across diverse applications~\cite{Bommasani2021OnTO}.
They hold promise for enhancing the full robotics stack, from perception to decision-making and control~\cite{firoozi2023foundation}.
Recent works integrate LLMs and VLMs as high-level planners paired with low-level skills to enable open-vocabulary and open-world robot capabilities~\cite{shah2024bumbleunifyingreasoningacting, Zhi2024ClosedLoopOM, Liang2022CodeAP, Hu2023DeployingAE, ahn2022i, Huang2022InnerME}.
Other studies use LLMs to model humans~\cite{Zhang2023LargeLM}, estimate uncertainty~\cite{ren2023robotsaskhelpuncertainty}, or use language~\cite{chang2023var, chang2023data, liu2024dragon}.
However, these approaches do not address the interpretation of user control inputs in real-world assistive teleoperation settings. Thus, the potential of LLMs/VLMs for assistive teleoperation remains underexplored.

\section{Assistive Teleoperation with \methodname{}}

\begin{figure*}[t]
    \centering
    \includegraphics[width=1\linewidth,trim=0 2 0 85,clip]{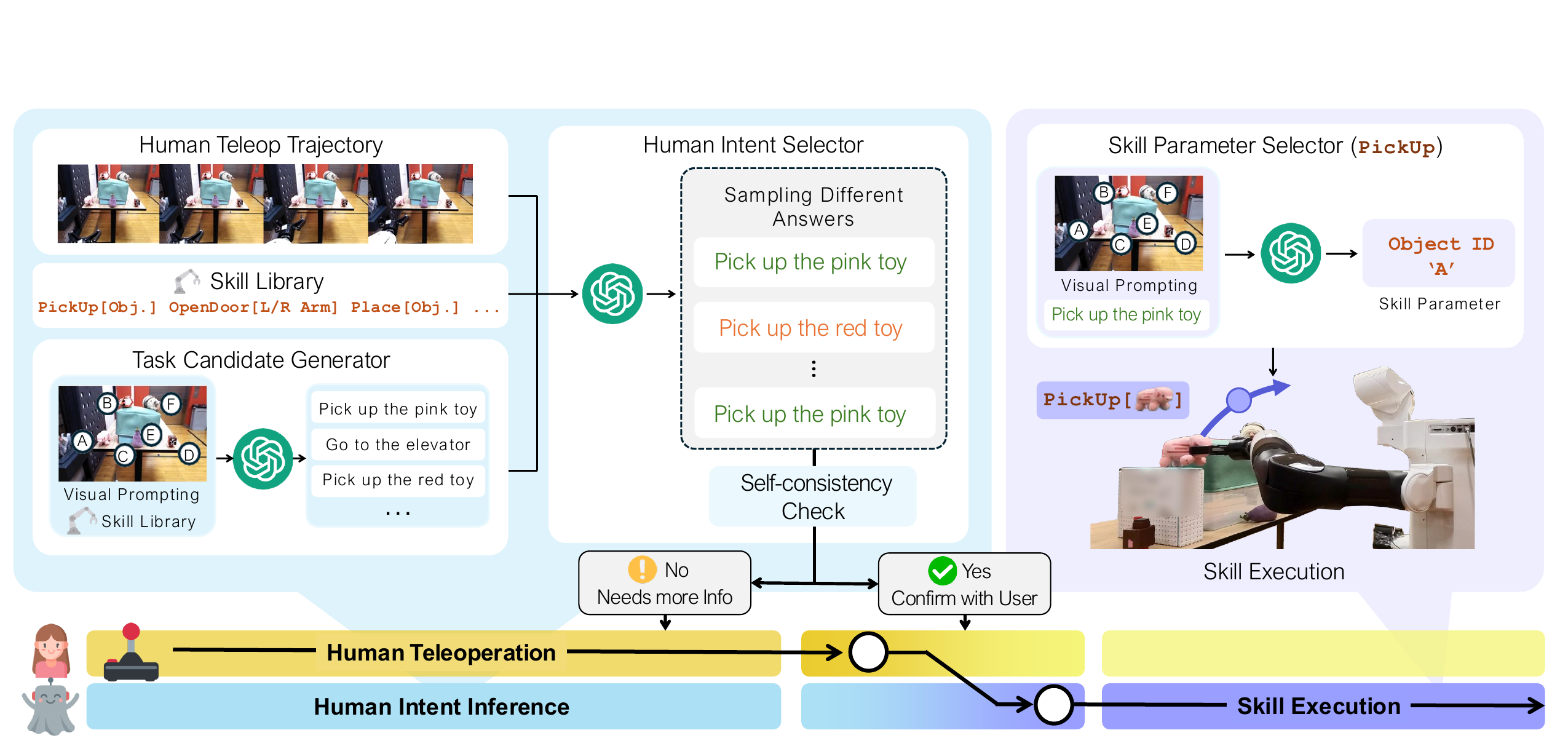}
    		\captionof{figure}{\textbf{\methodname{} architecture.} 
            VLM-based intent inference runs in parallel with human teleoperation. \methodname{} generates task candidates from observations and infers intent from user inputs among the task candidates, repeating until predictions are self-consistent. Once confirmed by the user, \methodname{} executes the corresponding skill with estimated parameters. 
            \label{fig:method}
            \vspace{-2mm}
            } 
\end{figure*}

In this section, we describe \methodname{}, an assistive teleoperation system that enables robots to infer and execute diverse human intents (Fig.~\ref{fig:method}). \methodname{} comprises two key components: an intent inference module that continuously predicts human intent from teleoperation history when shadowing the user in the background, and a skill execution module that executes tasks using a library of skills. 

\subsection{Problem Formulation}

We formulate assistive teleoperation as a sequential decision-making problem defined by the tuple $\langle \mathcal{S}, \mathcal{A}, \mathcal{P}, \mathcal{Z} \rangle$, where $\mathcal{S}$ is the state space, $\mathcal{A}$ is the action space, $\mathcal{P}: \mathcal{S} \times \mathcal{A} \rightarrow \mathcal{S}$ is the unobserved transition function, and $\mathcal{Z}$ is the intent space.
The state $s \in \mathcal{S}$ comprises the robot’s RGB image observation, proprioceptive states (e.g., gripper status, base and end-effector poses), and a list of foreground objects $O = \{o_1, ..., o_n\}$ detected from the open-world perception module.
The action $a \in \mathcal{A}$ is either from the human ($a = a_h$) during human teleoperation or from the robot ($a = a_r$) during autonomous execution. 
We assume that each assistive teleoperation episode is a sequence of one-step subtasks, and users can teleoperate to express their desired goals.
We define a human intent for the $i$-th subtask as $z_i = (l_z^i, o_z^i) \in \mathcal{Z}$, where $l_z^i \in L$ is the intended skill (e.g., ``navigate") and $o_z^i \in O$ is the target object (e.g., ``the door" in ``navigate to the door").
At the start of subtask $i$, the user provides a teleoperation trajectory snippet $\xi_h^T = (a_h^1, ..., a_h^T)$, where $T$ is the snippet length. 
The goals of \methodname{} are to infer the human intent $z_i$ from $\xi_h^T$, and to fulfill the intent with a trajectory $\xi_r$.
This process repeats until the human indicates the end of the episode.

\subsection{Inferring Intents in the Background}

\methodname{} tackles two key challenges in intent inference.
To identify intent, it generates possible candidates from open-world observations and selects the most likely one based on commonsense understanding of user inputs.
To handle intent ambiguity, it uses confidence estimation to predict only when confident, reducing premature suggestions.

\textbf{Intent Candidates Generation.} To generate an open set of potential intent options, we use a VLM $f_{candidate}$ to analyze the current state $s^t$ and create a set of intent candidates $\{c_1, ..., c_m\}$ (Fig.~\ref{fig:method} left).
It first identifies actionable objects and then filters feasible object-skill pairs based on how each object is likely to be interacted with.
The VLM adapts its predictions to object affordances and the robot’s current state (e.g., avoiding ``place" actions when the gripper is empty) by reasoning about robot-object interactions in a zero-shot manner.
The commonsense-based intent set generation ensures that intent choices are semantically plausible and relevant to the scene.

\textbf{Human Intent Selection.}
Given a set of task candidates $\{c_1, ..., c_m\}$, a second VLM $f_{intent}$ predicts the user's intent $\hat{z}$ by analyzing a history of subsampled robot observations, which include downsized images, robot base and end-effector poses (Fig.~\ref{fig:method} middle).
It chooses the most likely intent $\hat{z}$ among $\{c_1, ..., c_m\}$, and parse the corresponding skill class $l_{\hat{z}}$. 
To enhance VLM understanding in cluttered scenes, we apply visual prompting \cite{bahng2022exploringvisualpromptsadapting, shah2024bumbleunifyingreasoningacting, nasiriany2024pivotiterativevisualprompting} to annotate important regions that the VLM should attend to. These annotations include Set-of-Marks (SoM)~\cite{yang2023setofmarkpromptingunleashesextraordinary} for segmented objects, gripper masks that highlight gripper position, and arrows indicating gripper motion history.

\textbf{VLM Confidence Estimation.}  
Real-time intent inference is inherently uncertain due to the ambiguity or incompleteness of human actions. 
For instance, if a user begins rotating a robot's base in a room with multiple furniture pieces, the intended target remains ambiguous until the user clearly moves the robot toward a specific furniture. Seeking for user confirmation based on a premature guess can disrupt user control and cause frustration.
To address this, \methodname{} employs a confidence-based intent validation mechanism. Inspired by self-consistency methods \cite{wang2023selfconsistencyimproveschainthought} in LLMs, we run multiple VLM calls in parallel to estimate the confidence of intent predictions. The system only offers assistance when the number of VLM outputs in agreement exceeds a threshold. 
Formally, let $K$ denote the number of VLM calls and $\hat{z}^k$ the intent predicted by the $k$-th VLM. The system confirms its prediction with the user if 
$\sum_{k=1}^{K} \mathbb{I}(\hat{z}^k = \hat{z}^{\text{mode}}) \geq \eta$,
where \( \mathbb{I}(\cdot) \) is the indicator function, \( \hat{z}^{\text{mode}} \) is the most frequent prediction, and \( \eta \) is the agreement threshold.
By filtering out low-confidence predictions, this module minimizes disruptions and premature predictions.

\textbf{Parallel Foreground-Background System Design.}
Integrating pre-trained VLMs into real-time closed-loop control poses challenges due to the latency in VLM inference. Waiting for VLM outputs can be frustrating for users, especially when the system is uncertain or incorrect. To mitigate this delay, we adopt a framework where the user operates the robot in the foreground, while the VLM processes inputs simultaneously in the background. If the VLM is still processing or lacks confidence, it remains silent, intervening only when it has a confident prediction.
This approach allows the user to operate naturally while the system continuously refines its intent inference.

\subsection{Fulfilling Intents with Skill Execution}

Once confidence in its prediction, \methodname{} executes the intent using a library of parametrized skills, with a VLM estimating the skill parameters for execution.

\textbf{Control Switching.}
When confident in its prediction, the robot communicates the suggested action via an audible cue. The user can confirm or deny the prediction by pressing different keys on the keyboard. If confirmed, the system signals the transition to autonomous execution with another cue (``Great! I will take over.''). If denied, the system prompts the user to continue teleoperation (``Understood, I'll pause here. Feel free to continue.'') until the next prediction attempt.

\textbf{Parametrized Skill Library.} 
\label{sec:skill_library} In real-world assistive settings, users may require help with long-horizon tasks that involve diverse manipulation and navigation behaviors. 
\methodname{} utilizes a library of parameterized skills that cover common mobile manipulation behaviors, including object manipulation skills (e.g., picking, placing, pouring), interactions with the environment (e.g., pushing doors, tapping card readers, pressing buttons, taking elevators), and navigation (e.g., approaching landmarks).
Each skill is defined by a behavior primitive (e.g., \texttt{PickUp[Obj.]}) and a parameter (e.g., the target object's pose), enabling flexible execution of user intents across diverse environments. 
Refer to Appendix \ref{app:skill-lib} for a complete list of skills.

\textbf{Skill Parameter Selection and Execution.} Once a predicted intent $\hat{z}$ is confirmed (e.g., pouring pasta into a pot), the corresponding skill $l_{\hat{z}}$ (e.g., pouring) is called. The parameter estimation VLM $f_{skill}$ identifies parameters such as the target object $o_{\hat{z}}$. 
Based on the object's pose, the skill execution module executes the skill. 
After completing the subtask, the robot prompts the user to resume control (``Alright, you can take over now.”) for the next intent.

\begin{figure*}[t]
    \centering
      \includegraphics[width=0.95\linewidth]{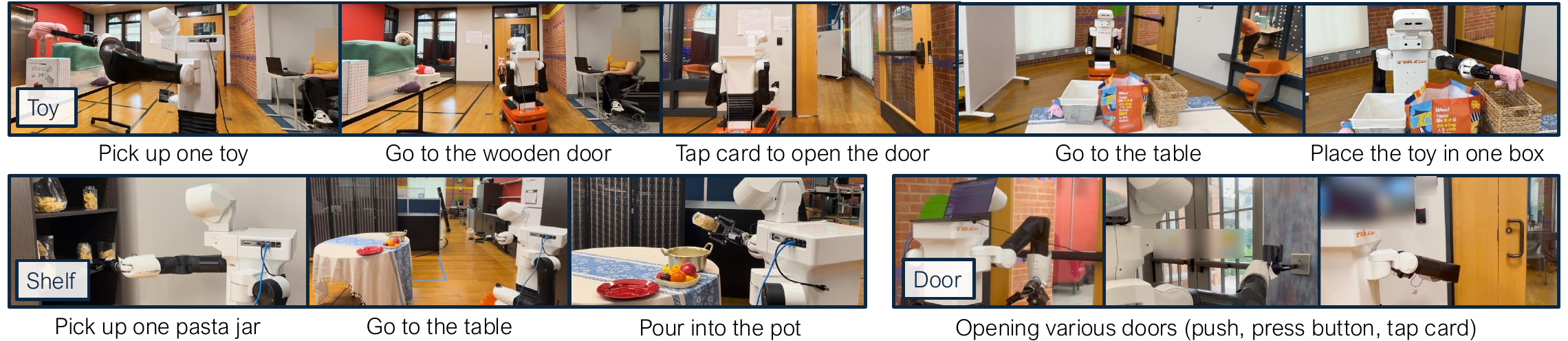}
		\captionof{figure}{\textbf{Toy, Shelf, and Door: multi-step mobile manipulation tasks.}
        At each step, the robot disambiguates user intent among multiple plausible goals, selecting the correct one based on user inputs and visual context. \label{fig:tasks}
        }
        \vspace{-3mm}
\end{figure*}

\section{Experiments}

We seek to answer the following research questions: \textbf{RQ1}: Does \methodname{} improve task performance and user experience compared to existing methods? \textbf{RQ2}: Is commonsense VLM reasoning essential for inferring diverse intents?
\textbf{RQ3}: What is the contribution of each system component to overall performance?
We address RQ1 through a user study, RQ2 via offline unit testing of the intent inference module, and RQ3 through ablation experiments.

\subsection{User Study: Real-World Mobile Manipulation Tasks}

\textbf{Experiment Setup}.
We use a TIAGo mobile manipulator equipped with dual arms, a mobile base, and an RGBD camera.
Users teleoperate the robot using a 3Dconnexion SpaceMouse while observing livestreamed RGB images.
\methodname{} uses GPT-4o as its VLM backbone.
The full teleoperation interface details, sensory setup, and audio/keyboard interaction design are provided in Appendix~\ref{app:exp-setup}.

\textbf{Tasks.}  We evaluate on $3$ tasks (Fig.~\ref{fig:tasks}) each requiring multi-step intent inference: \emph{Shelf} ($3$-step), 
\emph{Toy} ($5$-step), and \emph{Door} ($2$-step, $3$-variations). Each step offers multiple plausible choices, 
\begin{wraptable}[15]{r}{0.4\textwidth} %
\centering
\scriptsize
\setlength{\tabcolsep}{2.5pt}
\begin{tabular}{l S S S S}
\toprule
\multicolumn{5}{c}{\textbf{Task Success Rate (\% $\uparrow$)}} \\
\midrule
 & {Full Teleop} & {HAT} & {RBII} & {\textbf{\methodname{}}} \\
\midrule
Shelf   & 75.0 & 8.3 & 44.4 & \textbf{83.3} \\
Toy     & 79.2 & 37.5 & 33.3 & \textbf{91.2} \\
Door    & 75.0 & 75.0 & 57.1 & \textbf{91.2} \\
Average & 76.4 & 40.3 & 45.0 & \textbf{88.9} \\
\bottomrule
\end{tabular}
\vspace{2mm}

\begin{tabular}{l S S S S}
\toprule
\multicolumn{5}{c}{\textbf{Task Completion Time (s $\downarrow$)}} \\
\midrule
 & {Full Teleop} & {HAT} & {RBII} & {\textbf{\methodname{}}} \\
\midrule
Shelf   & 256.5 & 225.0 & 252.4 & \textbf{196.1} \\
Toy     & 391.2 & 406.3 & 388.3 & \textbf{362.5} \\
Door    & 120.4 & 112.6 & 118.7 & \textbf{96.8} \\
Average & 256.0 & 248.0 & 253.1 & \textbf{218.5} \\
\bottomrule
\end{tabular}
\caption{\textbf{User study: task success rate and completion time.} \methodname{} outperforms baselines in both task success and completion time.}
\label{tab:task_results}
\end{wraptable}
requiring the system to use user input to infer intents. More task details are in Appendix \ref{app:tasks}.

\textbf{Participants and Procedures.} 
We conducted an IRB-approved user study with $N=13$ participants (mean age $= 29.4$; $5$ females, $8$ males; all able-bodied)
, all of whom gave informed consent.
Participants completed a practice session before completing each method in randomized order. After each, they answered user satisfaction and NASA-TLX questionnaires.

\textbf{Independent Variables (Robot Control Methods).} 
We compare \methodname{} with three baselines: 1) \textit{Full Teleop}: The user manually teleoperates the robot without autonomous robot control. 2) \textit{HAT}~\cite{Padmanabha_2024}: assistive teleoperation  
that infers human intents using proximity to goal.
3) \textit{RBII}~\cite{argall2018autonomy}: assistive teleoperation that infers human intents using Bayesian inference using temporal user input history.
Since HAT and RBII only support grasping, we use \methodname{} to predict the skill and let the baselines select the target object, making comparisons conservative in their favor.
These baselines test the role of commonsense reasoning in diverse intent inference.

\begin{figure*}[t]
    \centering
    \includegraphics[width=1.0\linewidth]{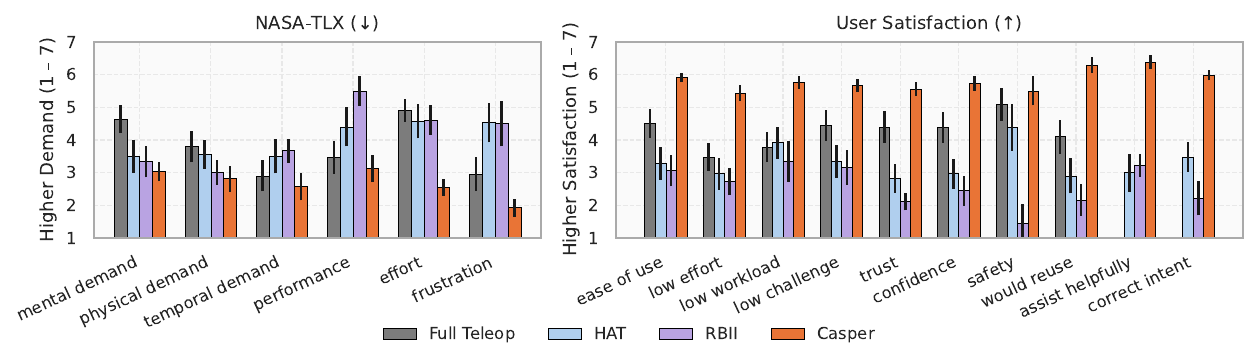}
		\captionof{figure}{\textbf{User study:  user workload and user satisfaction.} \methodname{} consistently outperforms the baselines in terms of user workload (left) and user satisfaction (right) with statistical significance ($p<0.05$). Detailed per-task results and full questions of user satisfaction can be found in Appendix \ref{app:user_study}. Note that for user satisfaction scores, ``assist helpfully'' and ``correct intent'' are not applicable to Full Teleop.}
        \label{fig:res_user}
        \vspace{-2mm}
\end{figure*}

\textbf{Dependent Measures (Evaluation Metrics).} 
To evaluate task performance, we measure the binary task success rate (completion in a fixed time limit). We measure human workload with NASA-TLX \cite{nasa_tlx_wikipedia}, a standard tool for evaluating subjective cognitive and physical workload. User satisfaction is measured with a questionnaire adapted from prior work \cite{Cui_2023}. We perform pairwise t-tests between \methodname{} with baselines to evaluate statistical significance.

\textbf{Hypotheses.} The user study tests the following hypothesis:

\begin{itemize}
    \item \textbf{H1}: \methodname{}'s VLM-driven intent inference and skill execution improve task performance over baselines in real-world assistive tasks;
    \item \textbf{H2}: \methodname{} reduces user workload and improves user satisfaction compared to baselines.
\end{itemize}

\textbf{Results.} 
\textit{Task Performance.} \methodname{} exhibits significant improvements ($p < 0.05$) in task success rate compared to all baselines (see Table \ref{tab:task_results}). 
The high success rate reflects the system’s ability to infer intents and execute appropriate actions, even in complex scenarios and long-horizon tasks.
Full Teleop is the runner-up in terms of success, allowing a portion of participants to succeed with expertise and patience. 
In contrast, \heuristic{} and \heuristicbayesian{} have lower success rates because they struggle with tasks requiring context or commonsense knowledge. 
We also report task completion time in Table \ref{tab:task_results}, where \methodname{} is the lowest across all tasks. Full Teleop is slower due to manual high-precision control (e.g., pouring); the heuristic baselines suffer frequent errors and corrections.

\textit{NASA-TLX.} Fig. \ref{fig:res_user} (\textbf{Left}) shows that \methodname{} significantly outperforms ($p < 0.05$) Full Teleop on all NASA-TLX metrics except ``performance", indicating that autonomous skill execution lowers cognitive and physical workload. Full Teleop requires continuous user input, resulting in higher workloads. 
The lack of statistical significance in ``performance'' suggests that user \textit{perceived} success is sensitive to \methodname{} occasional inference errors, despite \methodname{}'s \textit{objective} higher success rate.
\methodname{} also significantly outperforms ($p < 0.05$) \heuristic{} in all metrics except in ``mental demand" and ``physical demand," and significantly outperforms ($p < 0.05$) \heuristicbayesian{} across all metrics. The increased workload in \heuristic{} and \heuristicbayesian{} results from more frequent prediction errors (e.g., predicting to pick up the table), leading to longer time and higher effort. The results indicate that VLM-powered intent inference and skill execution reduce user burden and improve usability. The lack of statistical significance in ``mental demand" and ``physical demand" likely stems from assistive baselines sharing skill execution module with \methodname{}, which reduces the difference in these measures. 
\begin{figure*}[t]
    \centering
    \includegraphics[width=\linewidth]{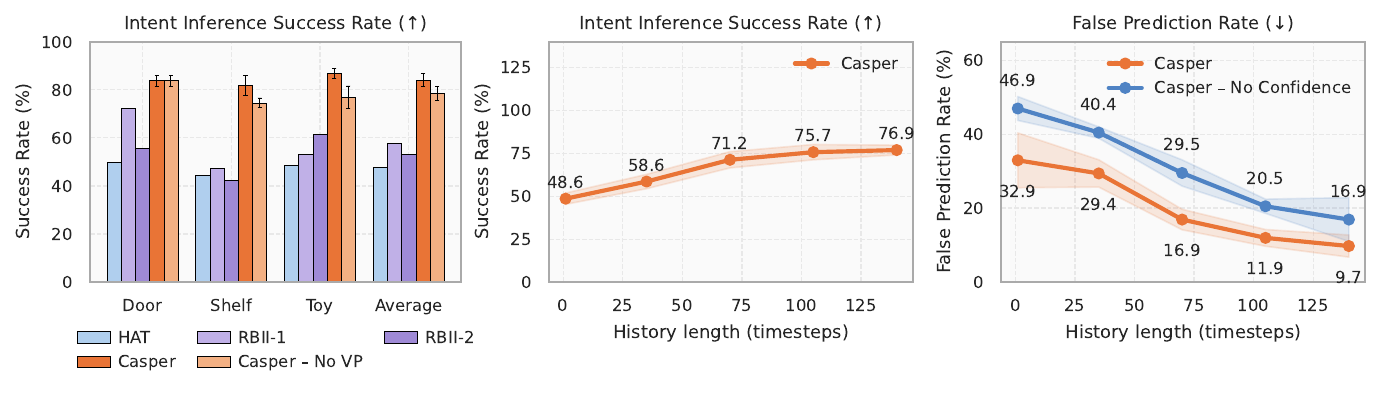}
    \vspace{-5mm}
		\captionof{figure}{
        \textbf{Quantitative results from unit testing and ablation studies.}
        \textbf{Left}: \methodname{} outperforms all baselines in intent inference success rate. Note that no STD is reported for deterministic baselines. The ablation of Casper vs. Capser - No Visual Prompting (VP) highlights the benefit of visual prompting. 
        \textbf{Middle}: Success rates improve with longer teleoperation history. 
        \textbf{Right}: Removing confidence estimation increases false prediction rates across all history lengths.
        \label{fig:res_offline}
        }  
        \vspace{-3mm} 
\end{figure*}

\textit{User Satisfaction.}
Fig. \ref{fig:res_user} (\textbf{Right}) shows that \methodname{} has statistically significant improvements ($p < 0.05$) in all $10$ user satisfaction metrics over all baselines. 
The results indicate that \methodname{} simplifies the assistance process and enhances the user experience. The \teleop{} baseline has lower scores, especially in ``effort'' and ``physical workload'' due to the demands of constant manual control. \heuristic{} and \heuristicbayesian{} score lower in ``confidence" and ``trust", as frequent intent prediction errors reduce user trust, significantly impacting overall user satisfaction.

In summary, the user study confirms that \methodname{} improves task performance (\textbf{H1}), reduces cognitive workload, and increases user satisfaction (\textbf{H2}). 
Beyond the main findings, the user study further reveals several notable insights which we detail in Appendix~\ref{app:user_study}, including more \textit{detailed analysis} of results, \textit{participant interviews}, a \textit{demographic breakdown}, and an analysis of \textit{failure cases}.%

\subsection{Unit Testing: Intent Inference Accuracy}

\begin{wrapfigure}{r}{0.2\textwidth}
  \centering
  \vspace{-10pt} %
  \includegraphics[width=0.2\textwidth]{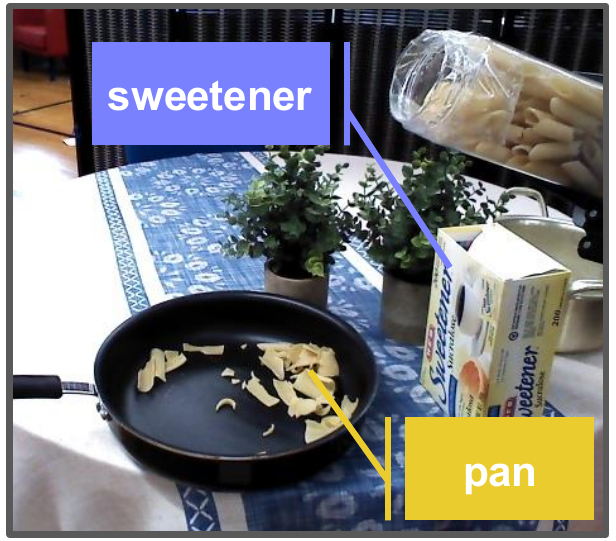}
  \caption{\textbf{Unit testing visualization.}}
  \label{fig:vis-heuristic}
  \vspace{-10pt} %
\end{wrapfigure}

To quantitatively validate \methodname{}'s intent inference accuracy, we conduct unit testing on teleoperation segments collected for each subtask across all three tasks.
Each segment serves as an independent data point for evaluating intent inference, where success requires correctly predicting both the intended skill and target object.
We prompt the VLM to predict the intended intent for each data point and compute the overall intent inference success rate, isolating intent inference from task execution.
In this experiment, we compare \methodname{} against the \heuristic{} and \heuristicbayesian{} baselines.
We evaluate two variants of \heuristicbayesian{} from the original paper: RBII-1's only uses the gripper-to-goal distance for recursive Bayesian inference, while RBII-2 also uses user joystick inputs with Boltzmann-rational action model.
Note that RBII-1 was used in the user study due to the similar average performance between RBII-1 and RBII-2.
As shown in Fig. \ref{fig:res_offline} \textbf{(Left)}, \methodname{} outperforms \heuristic{} and \heuristicbayesian{} baselines.
Without commonsense reasoning, the baselines often mispredict targets by relying on gripper motion trends toward nearby objects, e.g., incorrectly pouring pasta into a sweetener box (Fig. \ref{fig:vis-heuristic}) because the gripper moved closer to it.
In contrast, \methodname{}’s VLM-based inference leverages commonsense knowledge to make accurate predictions, choosing the pan instead.

\subsection{Ablation Studies}

To assess the impact of \methodname{}'s key components, we perform ablations on the following questions:

\textbf{How does the VLM input design, like visual prompting, affect intent inference accuracy?} To guide the VLM's attention more on user input changes and the manipulated object, we apply visual prompting (VP) by adding a gripper mask and an arrow of the robot gripper motion (rendered from proprioceptive states) on the image. Fig. \ref{fig:res_offline} \textbf{(Left)} shows VP yields an average $5.7\%$ boost, especially on Toy ($+9.8\%$). Removing VP hurts the success rate because the VLM must implicitly understand the robot end-effector movement trace.
Nonetheless, \methodname{}'s no-VP variant still outperforms all non-VLM baselines by $>10\%$, confirming that the primary gains come from the VLM’s commonsense reasoning; VP enhances its reasoning rather than providing decisive extra information. 

\textbf{How much human teleoperation history is needed for accurate intent inference?} 
\methodname{} infers intent from a segment of the user’s teleoperation trajectory. Short histories risk ambiguity and incorrect predictions, while long histories increase user effort. 
We investigate the tradeoff by studying the accuracy of intent inference across different trajectory lengths. We vary history length from $T=4$ to $T=140$ timesteps and measure intent inference accuracy, defined as correctly predicting both the skill and target object.
As shown in Fig.~\ref{fig:res_offline} \textbf{(Middle)}, longer histories improve accuracy by providing more context. However, gains plateau beyond $T=100$, offering diminishing returns while adding user burden. Thus, we use $T=100$ in the user study to balance accuracy and workload.

\begin{wrapfigure}{r}{0.6\textwidth}
    \centering
    \vspace{-10pt} %
    \includegraphics[width=0.6\textwidth]{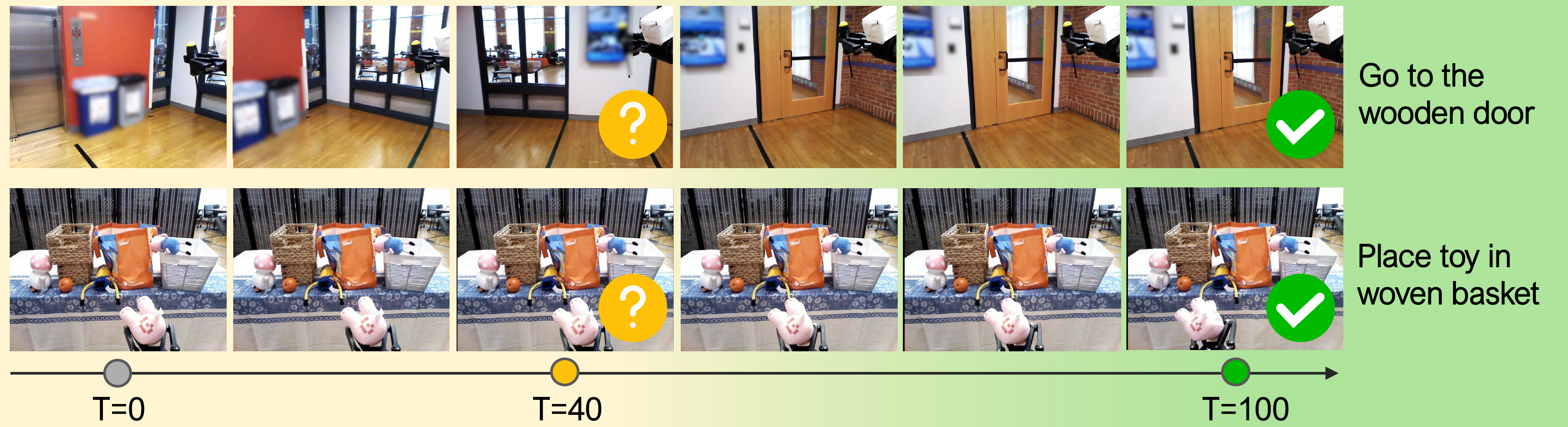}
    \caption{\textbf{Confidence estimation visualization.} \methodname{} predicts until the intent is clearer, ensuring more accurate assistance. }
    \label{fig:confidence}
    \vspace{-6pt} %
\end{wrapfigure}
\textbf{How does confidence estimation mitigate incorrect intent predictions?}
We hypothesize that \methodname{}’s confidence estimation module reduces false predictions by filtering out ambiguous cases. 
To validate this, we ablate the module and compare false prediction rates (defined as incorrect predictions over total predictions).
As shown in Fig.~\ref{fig:res_offline} \textbf{(Right)}, \methodname{} with uncertainty estimation consistently achieves lower false prediction rates, showing that uncertainty estimation effectively defers predictions when intents are unclear.
Fig.~\ref{fig:confidence} also shows qualitative examples. At $T = 40$, \methodname{} withholds predictions in both tasks due to ambiguity: The viewpoint is still shifting in the ``Go to the wooden floor" task, and the gripper movement is still unclear between the basket and bag in the ``Place the toy" task.
Premature inference could have led to incorrect predictions (e.g., selecting the wrong landmark or container). By $T = 100$, enough context enables correct predictions. These examples illustrate how delaying decisions in uncertain situations improves reliability.

\section{Conclusion}

We presented \methodname{}, an assistive teleoperation system that addresses the challenge of intent inference for mobile manipulators in real-world environments. \methodname{} interprets human intents from teleoperation inputs by leveraging the commonsense reasoning capabilities of pre-trained VLMs, and features an open-world perception module, a flexible library of parameterized skills, and a parallel inference-execution architecture. Extensive user studies and system evaluations demonstrate that \methodname{} outperforms both direct teleoperation and assistive baselines in success rate, user satisfaction, and mental workload.
Future work will explore continual learning of new skills from human interactions~\cite{grannen2024vocal,parakh2024lifelong} and improving intent inference reliability with uncertainty quantification techniques such as conformal prediction~\cite{ zhao2024conformalizedteleoperationconfidentlymapping, knowno2023}. %

\clearpage

\section{Limitations}
\methodname{} has the following limitations. 
First, our user study included operators of different ages, skill levels, and teleoperation experience to test \methodname{} under varied interaction styles. However, to further validate the system's benefits, a necessary next step is to involve users with motor or cognitive disabilities \cite{Padmanabha_2024, jenamani2024flairfeedinglonghorizonacquisition} and other underrepresented groups to cover a wider user distribution. Second, while the framework is compatible with the integration of continual learning of new skills \cite{grannen2024vocal, wan2024lotuscontinualimitationlearning} from user interaction, that capability is not yet included in the scope of this paper and is left for future work. Lastly, \methodname{} assumes that the user’s intent can be modeled with a combination of skills and target objects; finer-grained intents, such as precise motion paths, styles, or expressive behaviors \cite{Mahadevan_2024}, are not yet supported and will be pursued in future work.

\acknowledgments{We thank all participants in the human and pilot studies for their time and valuable contributions to our experiments. We thank Arpit Bahety, Gu-Cheol Jeong and Luca Macesanu for helping with Tiago hardware. We thank Ruta Desai, Roozbeh Mottaghi, Xavi Puig, Melanie Sclar and Changhao Wang for their fruitful discussions. This work was partially supported by the National Science Foundation (FRR-2145283, EFRI-2318065), the Office of Naval Research (N00014-24-1-2550), the DARPA TIAMAT program (HR0011-24-9-0428), and the Army Research Lab (W911NF-25-1-0065). It was also supported by the Institute of Information \& Communications Technology Planning \& Evaluation (IITP) grant funded by the Korean Government (MSIT) (No. RS-2024-00457882, National AI Research Lab Project). }

\bibliography{main}  %

\appendix

\clearpage

\section{Methods}

\subsection{Parametrized Skill Library.}
\label{app:skill-lib}

\methodname{} has $8$ parameterized skills, from navigation (\texttt{GoToLandmark}, \texttt{NavigateToPoint}) to object (\texttt{Pour}, \texttt{Pick}, \texttt{Place}) and environment interactions (\texttt{TapCard}, \texttt{PushDoor}, \texttt{PressButton}). The detailed skills and their descriptions are presented below:

\begin{lstlisting}
skill_name: pick_up_object
arguments: object_of_interest
description: pick_up_object skill moves its arms to pick up the object specified in the argument object_of_interest. The pick_up_object skill can only pick up objects within the reach of its arms and does not control the robot base.

skill_name: place_object
arguments: object_of_interest
description: place_object skill moves its arms to place what it is holding to the object specified in the argument object_of_interest. The place_object skill can only place objects within the reach of its arms and does not control the robot base. The robot should place objects onto containers like trash cans, sinks, or items with surfaces like chairs, tables, etc.

skill_name: tap_card_open_door
arguments: None
description: Opens the door by tapping the key card access, if key card access is needed.

skill_name: goto_landmark
arguments: Selected landmark image from the environment from various options.
description: Navigates to the landmark in the environment, example, bedroom, kitchen, tool shop, etc.

skill_name: navigate_to_point_on_ground
arguments: object
description: Moves the robot to a point near the selected object. This skill can be used to move to a point in the room to perform a task, example, navigating near the toaster to make a toast. You should use this skill if there is still some distance between the robot and the object of interest, e.g. there is some area on the floor in between, or there is some hallway. Do not use this if the robot is near and can reach the object directly.

skill_name: push_open_door
arguments: None
description: Opens the door if the robot is in front of the door. The robot moves forward to open the door.

skill_name: pour_object
arguments: object_of_interest
description: pour_object skill moves its arms to pour whatever it is holding to the object (containers) specified in the argument object_of_interest. The pour_object skill can only pour to objects (containers) within the reach of its arms and does not control the robot base.

skill_name: press_button
arguments: button position depending which button you want to press.
description: Equips the robot with the capability of pressing a button. The robot will push the button selected in the argument. The subtask must indicate which button to press, example, 'Press the accessibility button to open the door.'

\end{lstlisting}

\methodname{} uses open-world perception module with GSAM \cite{ren2024groundedsamassemblingopenworld} and GroundingDINO \cite{liu2024groundingdinomarryingdino}. Skills are parameterized on object pose. Low-level actions are then generated via Inverse Kinematics for joint configuration (arm) and motion planning on a 2D occupancy map (base). 
Adding a new skill is straightforward with a text description for VLM, a parameter mapping, and a few optional joint waypoints.  

\vspace{-3mm}

\subsection{Intent Inference Details.}

\textbf{VLM Prompts.} 

\textit{1) Prompt for Intent Candidates Generation}

\begin{participantquote}
    
First, give a list of possible tasks to perform, using the information of the scene, the relevant objects, and relevant skills. 

        Note:
        
        - The robot can only manipulate objects that are within 0.7 meters of the robot. If the distance from the robot to the object is greater than 0.7 meters, then you SHOULD NOT include the manipulation skills in the task choices!
        
        - The pick up and place skill should be used on smaller objects, and the navigate skill should be used on furniture, like tables, chairs, etc.
        You should use the robot history: Eliminate the tasks that the robot has already performed. If the robot has picked up an object, it will not perform the task again!

        Formulate your results in the format of multiple-choice questions.

        Example 1: Given that I am farther away and the robot is moving, the possible subtasks to perform are:
        
            A) Navigate to the desk with pens on top of it.
            
            B) Navigate to the brown colored door.

        Example 2: Given that I am near the table, the possible subtasks to perform are:
        
            A) Place the apple in the pink bowl.
            
            B) Pick up the screwdriver with blue handle.

        Example 3: Given that I am near the table, the possible subtasks to perform are:
        
            A) Pick up the blue bowl with pink stripes.
            
            B) Pick up the apple.
            
            D) Pick up the purple bowl.

        Example 4: Given that I am in the corridor, the possible subtasks to perform are:
        
            A) Go to the kitchen.
            
            B) Go to the classroom.

\end{participantquote}

\vspace{5mm}
\textit{2) Prompt for Human Intent Selection}

\begin{participantquote}

        INSTRUCTIONS:

        You are given a sequence of images of the scene. The images are taken from the camera on a mobile robot that is moving its base. Your goal is to determine the robot's intent based on this sequence of robot observations. You want to make use of the list of skills, the history of the robot's movement, and the list of task choices to determine the human's goal.

        The list of skills that the robot has are below. The tasks are using the skills listed here.
        \{skill description\}

        HISTORY OF PAST EXECUTIONS: You should make use of this information for decision making.

        \{history prompt\}

        Possible task choices:

        \{subtask prompt\}

        Think step by step, keep in mind the following points:

        1. Consider the given task choices.

        2. Focus on the images, and see if there is a change in robot's point of view; see how it is moving and changing its position, or if the gripper is getting closer to one of the objects, or turning towards one of the landmarks. 

        If the robot gripper is moving, see where the gripper (as masked in the image) is moving towards based on the green arrow, and use that to determine the task choice option.

        Then, given the images and the robot's movement, summarize the previous the robot's movement.

        3. Then, summarize the previous executions made by the robot and feedback received from the human or environment.

        Finally, answer: What is the robot trying to do? Choose from the list of possible task choices.

        Example reasoning 1: The robot is moving towards the left, where there is a table with a bowl on it. Since it has already picked up an object, it most likely wants to place the object on the bowl. However, the distance to the table is farther for the robot (greater than 0.7 meters) to place the object. We should first navigate to the table with a bowl on it.

        Example reasoning 2: The robot arm is moving closer towards the apple. The apple is already within the reach of the robot, that is, less than 0.7 meters. Therefore, it is likely that the robot will pick up the apple.

        Example reasoning 3: The robot is moving towards the bookshelf with a book in its hand. It is most likely trying to place the book on the bookshelf. However, the robot is far away from the bookshelf. We should first navigate to the bookshelf.

        Example reasoning 4: The robot is moving towards the table which has a book on it. The robot tried to pick up the book before, but it failed due to IK solver issues. Since the robot is far away from the table, we should first navigate to the table with a book on it using the navigate skill.

        Example reasoning 5: The robot is near the book shelf which has one thriller and one comedy book. The robot tried to pick up the comedy book but the human stopped it. It is likely that the robot will try to pick up the thriller book.

        Example reasoning 6: The robot is moving towards the bookshelf with a book in its hand. The robot tried to place the book on the book holder, but it failed due to IK solver issues. Since the robot is far away from the book holder, we should first navigate to the bookshelf using the navigate skill.

        Example reasoning 7: The robot is holding a bottle in its hand. Given that there are several cups and containers on the table, It is likely that the robot will pour the liquid from the bottle into one of the cups or containers.

        Provide the skill name in a valid JSON format. Your answer at the end in a valid JSON of this format: \{\{'subtask': '', 'skillname': ''\}\}
        
        Avoid using the object id in the final JSON response. Describe the object(s) involved in the sub-task instead of using the object id in the JSON response. This is very important.
        
        You should only choose from the list of task choices provided! This is very important.

        If the arm is moving, you should see where the arm GRIPPER TIPS is moving towards, and use that to determine the task choice!! The gripper tip consists of 2 pointly black parts at the end of the gripper, don't consider the white part.

        For example, if the gripper is mostly staying on the table level or below the table, then most likely the user is choosing objects on the table (i.e. the bottom row). Else if the gripper moves above the table and obstructs the objects on the table, then most likely the user is choosing something above the table on the top row.

        You should judge the physical distance between the robot and the object. You can tell that by checking if there is some area (like ground, floor) in between robot and the object. If the robot is far away from the object, it will most likely perform NAVIGATION. It is unlikely that the robot will do tasks that involve touching the object if the robot is far away from the object, e.g. pick up, place, pressing button, tapping card, etc.

        Pay attention to where the MASKED gripper is moving, and the direction of the arrow of the robot arm's movement! 
        
        The arrow means the direction of the gripper's movement. For example, if the arrow is pointing up and right, it means the gripper is moving to something on the top right.

        ANSWER: Let's think step by step.
    
\end{participantquote}

\section{Experiments Details}

\subsection{Details on Experiment Setup} \label{app:exp-setup}

We utilize the TIAGo mobile manipulator robot as our platform, which features dual robotic arms and a mobile base. The robot is equipped with the following sensors and components: 
(1) a 3Dconnexion SpaceMouse, serving as the teleoperation interface that enables users to convey their intents through manual control
(2) An RGB-D camera is mounted on the robot’s head to capture visual data, providing RGB image streams to both the \methodname{} system and the user interface.
(3) A laptop computer is connected to the robot to display the live RGB image feed to the user. The computer's speaker notifies users when autonomous operation becomes available, using audio prompts generated via the OpenAI text-to-speech API. When the system's intent prediction surpasses a predefined confidence threshold, the user confirms their intent using the computer’s keyboard.
We employ GPT-4o as the backbone of our visual-language model (VLM) due to its strong multimodal reasoning capabilities, though other VLM architectures may also be suitable.

\subsection{Tasks} \label{app:tasks}

The three tasks in Fig. \ref{fig:tasks} are assistive mobile manipulation tasks in a public building, each requiring multi-step intent inference:

\begin{itemize}[left=0pt]
    \item \textit{Shelf} task has three subtasks: Picking up a pasta jar of the user's choice from multiple ones on a multi-layer shelf, navigating to a table, and pouring the pasta into a container selected by the user among multiple containers.
    \item \textit{Toy} task has five subtasks: Picking up a toy of the user's choice among multiple ones from a table, navigating to a door among multiple exits, opening the door by tapping the card on the card reader, navigating to a table among multiple pieces of furniture, and place the toy in one of the containers on the table. This task is especially challenging since it is a long-horizon task involving five substeps.
    \item \textit{Door} task has two subtasks: Navigating to a specified door among different landmarks and opening it using a method of the user's choice. This task consists of a suite of three different doors, each of which requires different ways of being opened. The task includes three door types: (1) push to open, (2) press an accessibility button or push to open, (3) tap an access card or pull the handle.
    
\end{itemize}
Our experiments span across $20+$ objects, $7$ furniture types, and $5$ rooms. The diverse objects and skills required in these tasks closely mimic the open-ended possibilities that assistive robots encounter in the real world. Furthermore, the robot must utilize human input to choose the correct action among multiple options in order to successfully complete the task. 

\section{More Human Study Results} 
\label{app:user_study}

\begin{figure*}[h!]
\centering
\vspace{2mm}
\includegraphics[width=\linewidth]{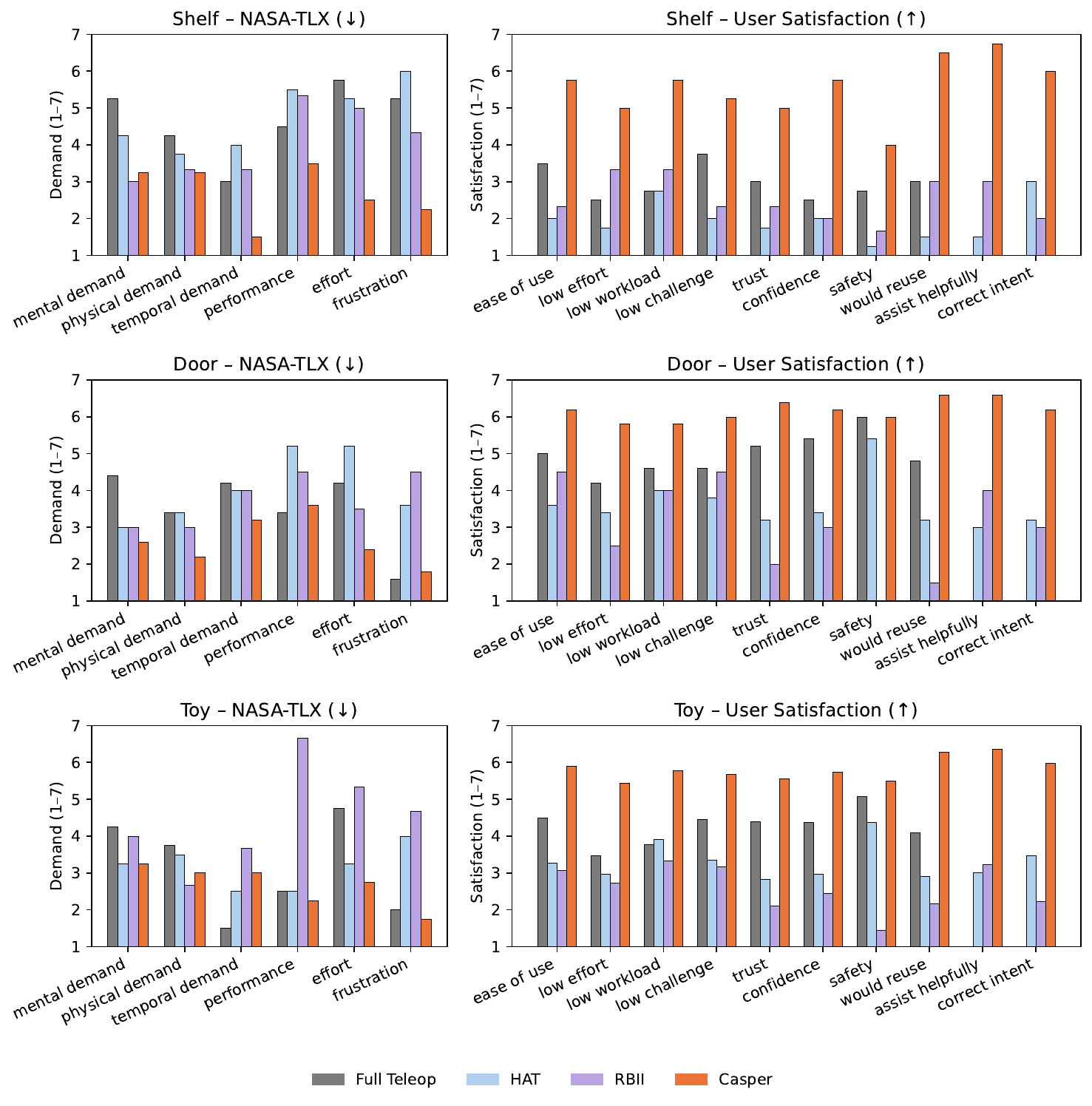}
\caption{\textbf{Per-task results for NASA-TLX and user satisfaction scores.} Note that for user satisfaction scores, ``assist helpfully'' and ``correct intent'' are not applicable to Full Teleop.}
\label{fig:per-task-results}
\end{figure*}

\textbf{Detailed Human Study Results.} We present per-task results for NASA-TLX and user satisfaction scores in Figure \ref{fig:per-task-results}.

\textbf{Participant Interview: Qualitative Feedback on Methods}. 
We present the qualitative responses from users to gain deeper insights into their experiences with each method. The feedback highlights key aspects of usability, intent inference accuracy, execution reliability, and user workload. Users generally favored \methodname{} for its ease of use and high success rate, while \teleop{} was perceived as slow and effort-intensive due to full manual control. \heuristic{}, despite attempting automated assistance, was often unreliable in execution, leading to frustration. These qualitative findings complement our quantitative results, reinforcing the importance of accurate intent inference, seamless execution, and minimizing user workload in assistive teleoperation systems.

\textit{1) \methodname{}: most preferred, easiest to use, and most reliable}

Users generally praised \methodname{} for accuracy and low workload.

\begin{participantquote}
``... made it easiest to get the job done and I felt like I had the most success."

``... is very comfortable."

``Never misunderstand what I want to do and never make mistakes."
\end{participantquote}

Compared to other methods, \methodname{} required the least manual effort and had the best success rate in both intent inference and execution. However, some users mentioned that fine-grained control (e.g., pressing buttons) of the skill library could still be improved:

\begin{participantquote}
``The fine-grain control (e.g., press the door open, tap card) is not as smooth."
\end{participantquote}

\vspace{3mm}
\textit{2) \teleop{}: precise but demanding, slow, and effort-intensive}

\teleop{} was seen as high-effort and slow due to its reliance on full teleoperation, making it the most mentally and physically demanding. Users found it precise but burdensome, requiring them to control every aspect of the robot manually:

\begin{participantquote}
``I need to do all the control work by myself for [\teleop{}]."

`` ... takes a lot of human effort and is overall slowest to operate."

``... was annoying to microcontrol the robot but it worked fine."
\end{participantquote}

\vspace{2mm}
Although some appreciated its precision, they found it exhausting and inefficient, often resulting in slow operation:

\begin{participantquote}
``By self-controlling, the robot moves very slow."

``... is more precise but demanding."
\end{participantquote}

Overall, users acknowledged the controllability it offered but disliked the high workload.

\vspace{3mm}
\textit{3) \heuristic{}: unreliable, sometimes alarming, high failure rate}

\heuristic{} had mixed to negative feedback, with users describing it as unreliable, inconsistent, and prone to errors in both prediction and execution. While some users found its intent inference acceptable, execution failures made it frustrating:

\begin{participantquote}
``... was a bit alarming."

``Every time it will make some mistakes or misunderstand what I want to do."
\end{participantquote}

Failures in execution had a particularly strong negative impact on user perception, as they felt that errors led to frustration and loss of trust in the system:

\begin{participantquote}
"[\heuristic{}] was terrible."
\end{participantquote}

Despite some users acknowledging its attempt at autonomous assistance, they preferred not to use it again due to its inconsistency:

\begin{participantquote}
``[\methodname{}] is the best, [\teleop{}] is the worst. [\heuristic{}] is ok, but I don’t want to try again."
\end{participantquote}

\vspace{3mm}
\textit{4) \heuristicbayesian{}: unpredictable, imprecise, and occasionally concerning}

\heuristicbayesian{} also received mixed to negative feedback, primarily due to its unreliable intent inference and unpredictable behavior. Users frequently expressed concerns about safety and control transparency:

\begin{participantquote}
``The inferred intent is always wrong, and the control over the robot actions are always weird and potentially have security issues ... Would it accelerate towards me and hit me?"
\end{participantquote}

While some users acknowledged partial task completion, they noted that it was inconsistent and often required intervention:

\begin{participantquote}
``It was not that bad, it was in the midrange. I was able to complete one task completely and the other one was with difficulties."

``It does try to do the action, but it performs the wrong actions (e.g., press the wrong spot)."
\end{participantquote}

Overall, while \heuristicbayesian{} showed some capability, users found it untrustworthy and ineffective for complex tasks.

\vspace{3mm}
\textbf{Participant Interview: Challenges and Failure Cases of \methodname{}}. 
While \methodname{} was generally well-received for its accuracy and assistance, users highlighted areas for improvement, mainly focusing on execution reliability, responsiveness, and interaction timing. The key points for improvement include:

\textit{1) Execution failures at edge cases}

Some users noted that while the robot correctly inferred their intent, it occasionally failed to execute the task properly, making it harder to recover control:

\begin{participantquote}
    ``Although the robot knew what I was going to do, it sometimes failed in finishing the task and led to a state where it was harder to control the robot."
    
    ``Cannot aim at a specific object well."
\end{participantquote}

\vspace{3mm}
\textit{2) Slow responsiveness and speed}

Several users felt that \methodname{} was too slow, particularly when transitioning between teleoperation and autonomous assistance:

\begin{participantquote}
``It was a bit slow while I was in control."

``Waiting for the robot to respond and move because it's slow."

``The inference for pressing the door open is not very prompt."
\end{participantquote}
\vspace{3mm}

\textbf{More on Human Study Protocol and Design.} We conduct a pilot study of group size $= 8$ before the official human study. The pilot study phase aims to evaluate the experimental setup's feasibility, refine task instructions, and identify potential usability issues. Specifically, we aim to assess system usability by ensuring that participants can effectively interact with \methodname{} and verify that the control of the system is intuitive. Second, we validate intent inference performance by analyzing whether the system reliably predicts user intent on a wide distribution of users and whether adjustments to inference thresholds or timing are necessary. We also gather preliminary user feedback by collecting qualitative insights on user experience, cognitive load, and overall satisfaction to inform improvements before the complete study.

\end{document}